\documentclass[runningheads]{llncs}

\usepackage[utf8]{inputenc}
\usepackage{amsmath}
\usepackage{float}
\usepackage{booktabs}
\usepackage{graphicx}
\usepackage{comment}
\usepackage{xcolor}
\usepackage{appendix}
\usepackage{hyperref}

\usepackage{amsmath}

\begin{document}
\title{AHAM: Adapt, Help, Ask, Model - Harvesting LLMs for literature mining}
\titlerunning{AHAM: Adapt, Help, Ask, Model}
\author{Boshko Koloski \inst{1,2} \and Nada Lavra\v{c}\inst{1,3} \and  Bojan Cestnik \inst{1,4}
  \\ 
Senja Pollak\inst{1}\and Bla\v{z} \v{S}krlj  \inst{1}\and 
Andrej Kastrin\inst{5}}
\authorrunning{Koloski et al.}
%
\institute{Jo\v{z}ef Stefan Institute, Ljubljana, Slovenia \\
\and International Postgraduate School Jo\v{z}ef Stefan, Ljubljana, Slovenia \\
\and University of Nova Gorica, Vipava, Slovenia  \\
\and Temida d.o.o, Ljubljana, Slovenia \\
\and
University of Ljubljana, Institute for Biostatistics and Medical Informatics, Ljubljana, Slovenia \\
\email{boshko.koloski@ijs.si, andrej.kastrin@mf.uni-lj.si}
}
\maketitle               
\begin{abstract}
In an era marked by a rapid increase in scientific publications, researchers grapple with the challenge of keeping pace with field-specific advances. We present the `AHAM' methodology and a metric that guides the domain-specific \textbf{adapt}ation of the BERTopic topic modeling framework to improve scientific text analysis.  By utilizing the LLaMa2 generative language model, we generate topic definitions via one-shot learning by crafting prompts with the \textbf{help} of domain experts to guide the LLM for literature mining by \textbf{asking} it to model the topic names. For inter-topic similarity evaluation, we leverage metrics from language generation and translation processes to assess lexical and semantic similarity of the generated topics. Our system aims to reduce both the ratio of outlier topics to the total number of topics and the similarity between topic definitions. The methodology has been assessed on a newly gathered corpus of scientific papers on literature-based discovery. Through rigorous evaluation by domain experts, AHAM has been validated as effective in uncovering intriguing and novel insights within broad research areas. We explore the impact of domain adaptation of sentence-transformers for the task of topic \textbf{model}ing using two datasets, each specialized to specific scientific domains within arXiv and medarxiv. We evaluate the impact of data size, the niche of adaptation, and the importance of domain adaptation. Our results suggest a strong interaction between domain adaptation and topic modeling precision in terms of outliers and topic definitions. 

\end{abstract}

\section{Introduction}

The large number of publications that appear every day makes it almost impossible for researchers to keep up with the latest findings, even within their narrow scientific field. In a flood of information, researchers can overlook valuable segments of knowledge. Text mining is an effective technology that not only supports the analysis of existing knowledge but also allows researchers to infer new knowledge facts based on information hidden in the data. Common text-mining tasks include information extraction from literature, document summarization, question-answering, and topic modeling, to name just a few. Topic modeling is a text-mining technique that enables researchers to explore the thematic landscape of a collection of documents from a high-level perspective~\cite{vayansky2020review}. Although topic modeling is an established research field, the recent advent of Large Language Models (LLMs) has led to a qualitative leap in their development.

This paper presents methodological advancements in domain adaptation for topic modeling by utilizing state-of-the-art language models. The approach is demonstrated in a corpus of scientific papers from literature-based discovery (LBD), a research field that our team has been working on for the last two decades~\cite{lavrac2020bisociative,kastrin2021scientometric}. The specific contributions of this paper are as follows:
\begin{itemize}
    \item \texttt{AHAM} - a topic modeling objective function that evaluates the quality of topic modeling by measuring, on the one side, the semantic and lexical similarity in generated topic names, and outliers on the other side, while adapting to a new domain.
        \item Two specialized sentence transformers fine-tuned for distinct arXiv domains: one for general science and another for biomedical science.
    \item A qualitative and quantitative assessment of domain-adaptation effects on topic modeling for literature-based discovery (LBD).
\end{itemize}

The remainder of this paper is organized as follows: Section \ref{sec:rw} discusses related work, Section \ref{sec:data} describes the data used, Section \ref{sec:method} outlines the proposed methodology, Section \ref{sec:experimetnal} explains the research questions and presents the results, and Section \ref{sec:fw} concludes with a qualitative and quantitative summary and discusses the further work.

\section{Related work}%
\label{sec:rw}

\par The field of natural language processing has witnessed a remarkable transformation with the advent of \textbf{Large Language Models} (LLMs), which can be divided into two groups: Masked Language Models (MLMs) like BERT \cite{devlin-etal-2019-bert} and generative Causal Language Models (CLMs) such as LLaMa2 \cite{touvron2023llama}. These foundational models have set new standards in understanding and generating text with human-level precision \cite{min2023recent}. Building upon this groundwork, sentence transformers \cite{reimers-2019-sentence-bert} have emerged as a specialized evolution. These models, tailored for the task of learning sentence representations, ensure that semantically similar sentences are closely aligned in the vector space. One notable application of sentence transformers is BERTopic \cite{grootendorst2022bertopic}, which has revolutionized topic modeling with its unique approach. BERTopic clusters sentences based on semantic similarity, providing a refined and context-sensitive thematic analysis that outperforms conventional methods. In a similar vein, KeyBERT \cite{grootendorst2020keybert} advances the field of keyword extraction. Utilizing sentence-transformer technologies, it effectively extracts key terms and phrases from extensive texts \cite{10.1007/978-3-031-18840-4_27,koloski-etal-2022-thin}. A pivotal area of research in the use of these models is domain adaptation. Wang et al. \cite{wang-2021-TSDAE} proposed an approach for unsupervised domain adaptation, employing sequential denoising auto-encoders to learn from corrupted data. Another approach to domain adaptation involves generative pseudo-labeling (GPL) \cite{wang-2021-GPL}, where researchers use a surrogate generative model, such as T5 trained to generate queries for specific passages \cite{thakur2021beir}. These queries are then ranked by a cross-encoder \cite{reimers-2020-multilingual-sentence-bert} and used as downstream fine-tuning data for the sentence transformer. The development of prompting techniques in LLMs \cite{wei2022chain}, particularly in-context one-shot learning \cite{lampinen-etal-2022-language}, represents a significant stride in model interaction. This approach involves crafting specific prompts that enable models to learn from a single example within the prompt context, thereby generating more relevant and contextually nuanced responses. This technique is crucial in eliciting accurate and specific outputs from models like LLaMa2 \cite{touvron2023llama}, demonstrating an advanced level of understanding and flexibility in language generation \cite{pan-etal-2023-context}.

\textbf{Literature-based discovery} (LBD) is a vibrant area of research, with the first approach reported in the mid-1980s. Swanson~\cite{swanson1986fish} observed, by reading separate literature on fish oil and Raynaud's disease, that some knowledge concepts are common between both document sets. This serendipitous discovery led Swanson to propose that fish oil may be used in the treatment of Raynaud's disease. Swanson's hypothesis was later clinically confirmed. 
\par LBD postulates that knowledge in one domain may be related to knowledge in another domain, but without the relationship being explicit. The general idea of LBD can be operationalized using three knowledge concepts, namely A, B, and C. If concept \( B \) is associated with concept \( A \) in the first document set, and concept \( B \) is associated with concept \( C \) in a disjoint document set, we may hypothesize a transitive relationship between concepts \( A \) and \( C \) through concept \( B \), which is common in both literature sets. 
\par Over the past four decades, drawing inspiration from the pioneering work of Swanson, numerous approaches to LBD have been developed. Researchers have proposed several different approaches, ranging from basic latent semantic indexing~\cite{gordon1998using} to techniques based on knowledge graphs~\cite{sang2018sematyp} and state-of-the-art large language models~\cite{wang2023learning}. An in-depth overview of LBD approaches can be found in recent studies~\cite{sebastian2017emerging,thilakaratne2019systematic}.

\section{LBD corpus and initial data analysis}
\label{sec:data}

We have compiled a comprehensive corpus of LBD publications by merging lists of representative publications on LBD that were manually prepared by the authors of two recent surveys of the LBD field~\cite{thilakaratne2019systematic,kastrin2021scientometric}. To ensure that the latest advances are included, we also added 11 papers published in the last two years that were not included in the previous surveys. Our corpus comprises 389 publications spanning from 1986 to 2023. We concatenated the title and abstract fields to eliminate empty features because six papers did not contain an abstract.

\begin{figure}[hbt]
   \centering
    \includegraphics[width=0.7\textwidth,height=5cm]{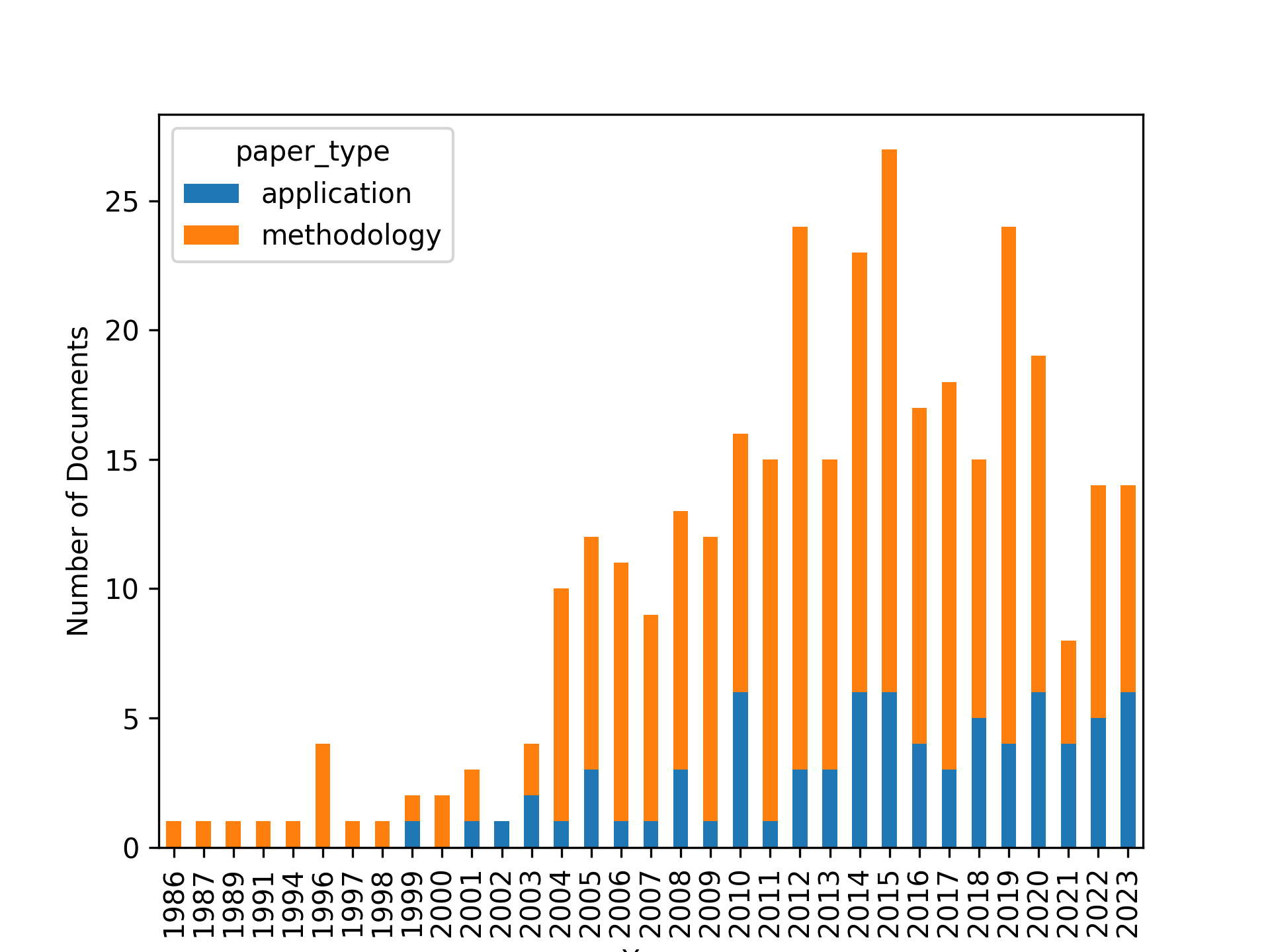} 
    \caption{Documents distribution per year, semantically labeled by LLaMa2 prompting. }
    \label{fig:sem-label}
\end{figure}

\par The publication frequency of these articles peaked in 2015, with variations observed over the years. We applied the LLama2 model to semantically categorize these articles into two groups: those proposing `methodology' and those applying already developed methodologies `application'. We utilize the one-shot learning capability of the LLama2 model. Our goal in this pseudo-categorization was to trace the evolution of the field, assuming that the `methodology' articles are those that introduce new techniques, while the `application' articles are likely to focus on the use or refinement of existing methods. We report the results in Figure~\ref{fig:sem-label}. The field of LBD started in 1986 with a paper reporting Swanson's~\cite{swanson1986fish} discovery of a link between fish oil and Raynaud disease, here categorized as a `methodology' paper. In the early years, the literature mostly explored methodological issues. Over time, the focus of work expanded to include both `methodology' and `application', with the latter gaining traction from 1999 onwards. A notable surge in `methodology' articles occurred around 2008, with a peak in 2015, followed by some fluctuation. On the other hand, `application' articles saw a steady increase, underscoring the sustained interest and necessity for LBD research. By 2023, the production of both `methodology' and `application' articles appears to have stabilized, suggesting a balanced advancement in these research domains.

\section{Methodology}
\label{sec:method}
In this section, we outline the proposed methodology, which builds on the versatility of the BERTopic framework and introduce the \texttt{AHAM} heuristic. 

\subsection{Corpus vectorization with Sentence Transformers}
\label{sec:st}
Sentence transformers \cite{reimers-2019-sentence-bert} optimize a loss function that measures the discrepancy between predicted and actual outcomes. A commonly used loss function for these transformers is either contrastive loss or triplet loss. Given a query sentence \( q \), a positive sentence \( p^{+} \) (semantically similar to \( q \)), and a negative sentence \( p^{-} \) (semantically different from \( q \)), we define the loss function as:
$ L(q, p, n) = \max(0, f(q, p^{+}) - f(q, p^{-}) + \text{margin}) $. Here, \( f(q, b) \) measures the similarity between sentences \( q \) and \( b \), often using cosine similarity, and the \(\text{margin}\) is a hyperparameter that establishes the minimum distance between the positive and negative pairs.

\subsection{Domain Adaptation via Generative Pseudo Labeling}
\label{sec:domain_adapt}
Given a sentence transformer embedding $S$ trained on a  source domain data $D_s$ and target domain data $D_t$, the goal is to adapt the sentence transformer embedding $S$ to the target domain data. One approach to do this is via generative pseudo-labeling (GPL) \cite{wang-2021-GPL}. The GPL adaptation involves multiple steps:  \textbf{Query Generation with T5}: To enable supervised training, first a generative model (e.g., T5), is employed to generate synthetic query sentences $Q_{synth}$ from the passages of the $D_t$, $Q_{synth} = \text{T5}(D_s)$. We generate up to $q$ queries per passage. \textbf{Mining Negative Passages}: For each generated query $q_i$ in $Q_{synth}$, negative passages $P_{neg}$ are mined using a pre-trained sentence embedding model. The sentences are mined via nucleus sampling to ensure diversity. In this manner, for each query $q_i$, the positive sample passage $p_i^{+}$ and the negative sample $p_i^{-}$ form a train tiplets $t$ = ($q_i$, $p_i^{-}$, $p_i^{+}$). \textbf{Scoring with Cross Encoder}: Next for each query triplet, a cross encoder $CE$ is utilized to score pairs of $q_i, p_i^{+}$ and $q_i, p_i^{-}$ as: $\delta_{(t_{i})} = CE(q_i, p_i^{+})) - CE(q_i,p_i^{-}) $. In this manner, we obtain a synthetic dataset adapted to the $D_{t}$ for fine-tuning the sentence-transformer $S$. Finally, the model $S$ is fitted to the new domain via MarginMSE loss-function \cite{DBLP:journals/corr/abs-2010-02666}. The final step of the vectorization phase projects the adapted sentence-transformer space into a lower dimension using UMAP \cite{McInnes2018} for dimensionality reduction. This step aims to make the clustering methods more effective, which otherwise struggle with high dimensional spaces. We configured the UMAP model with 5 nearest neighbors and 5 components, set the minimum distance to 0.0, used cosine similarity as the metric, and set a random state of 42 for reproducibility. In the subsequent phase of the BERTopic framework, we focus on identifying document clusters for later topic mining. We used an HDBSCAN  \cite{mcinnes2017hdbscan} model, set the minimum cluster size to 5, employed the Euclidean metric for distance calculation, chose the `eom' method for cluster selection, and enabled prediction data. Next, BERTopic calculates frequencies at the corpus level and creates a bag-of-words representation at the cluster level instead of for individual documents. This method highlights the importance of words at the topic level (i.e. the cluster level) and uses L1-normalization to accommodate clusters of different sizes, thus avoiding assumptions about the structural composition of the clusters. Finally, for each cluster, we employed the KeyBERT keyword extractor, which is based on the $S$ sentence-transformer representation, and subsequently ranks and retrieves the keywords. We retrieved a maximum of 10 relevant keywords per article and then ranked them. 
\par In the final stage of the processing pipeline for a set of $N$ documents, we obtain $k$ distinct clusters. Among these, one unique cluster is labeled as $k_{\text{outliers}}$, which represents the outliers. These outliers are the documents that did not fit into any of the other clusters.

\subsection{Prompt Engineering of LLMs to Design Topic Names}

\subsubsection*{Prompt engineering.}
We utilize the Llama2 language model, specifically the \emph{Llama-2-13b-chat-hf} variant from the HuggingFace\footnote{\url{https://huggingface.co/}}
repository, setting the temperature to $0$, the context window to $500$ tokens, and the repetition penalty to 1.1. The objective of this phase is to semantically utilize the extracted keywords and the most central documents in each topical cluster. We aim to leverage the LLM's capabilities to derive meaningful semantic labels. This is achieved through training via one-example in-context learning, guided by prompt designs crafted by domain experts of the meta-literature analysis application of our interest. Following related work, we have engineered a three-level prompt structure:
\begin{itemize}

    \item System prompt, to give personality to the LLM for labeling topics as a domain expert: \\
        \fboxsep=5pt 
        \fboxrule=1pt 
        \fcolorbox{black}{lightgray}{%
            \small 
            \parbox{11.25cm}{ 
            You are a helpful, respectful, and honest research assistant 
            for labeling topics.
            }%
        } \\

\item  One shot, example prompt, from which we \textbf{help} the model with guidance for in-context learning. Notice that this prompt has been crafted in a specific manner, reflecting our contributions to the field, but it could have been crafted differently by the end user.

        \fcolorbox{black}{lightgray}{%
            \small 
            \parbox{11.25cm}{ 
             I have a topic that contains the following documents: \\
            - Bisociative Knowledge Discovery by Literature Outlier Detection. \\
            - Evaluating Outliers for Cross-Context Link Discovery. \\
            - Exploring the Power of Outliers for Cross-Domain Literature Mining. \\    
            The topic is described by the following keywords: 
            bisociative, knowledge discovery, outlier detection, 
            ,data mining, cross-context, link discovery, 
            cross-domain, machine learning'. \\
            Based on the information about the topic above, 
            please create a simple, short, and concise computer science 
            label for this topic.     
            Make sure you only return the label and nothing more. \\
            $[$INST$]$: Outlier-based knowledge discovery
            }%
        } \\

    \item The query prompt, which the model uses to label topics and keywords of the most central documents for a given query, is personalized. This personalization ensures that the model returns computer-science topics from the field of Literature-Based Discovery without explicitly mentioning them: \\

        \fcolorbox{black}{lightgray}{%
            \small 
            \parbox{11.25cm}{ 
            I have a topic that contains the following documents $[$DOCUMENTS$]$ \\
            The topic is described by the following keywords: $[$KEYWORDS$]$ \\
            Based on the information about the topic above, 
            please create a simple, short and concise computer science 
            label for this topic. 
            Make sure you only return the label and nothing more.
            }%
        }     
\end{itemize}

\subsubsection*{Assessing Adaptation Through Evaluation of Topic Naming.}
To assess the effect of domain adaptation on topic modeling, we used the idea that, on average, the names of topic labels tend to be dissimilar. To measure the numerical similarities between topics, we have chosen to use three distinct similarity metrics \texttt{TopicSimilarity} to evaluate the similarity between topic names of Topic A and Topic B:
\begin{itemize}
    \item \textbf{Levenstein Fuzzy Matching}:  We employ the \emph{fuzzywizzy} python implementation which utilizes normalization based on the length to provide a normalized \textit{lexical} similarity score between the topic names.

    \item \textbf{BERTscore}: Uses BERT model embeddings to evaluate the similarity between A and B, by comparing the semantic embedding similarity of present n-grams \textit{w}
    $ \text{BERTscore} = \frac{1}{|A|} \sum_{w \in \text{A}} \max_{w' \in \text{B}} \text{cos}(w, w') $ 
    \item \textbf{Semantic Similarity (using all-mini-LM-v12)}: Evaluates the semantic closeness of A and B using the cosine of the angle between their vector representations, S(A) and S(B), as:
$ \text{cos(A, B)} = \frac{S(A) \cdot S(B)}{\|S(A)\| \cdot \|S(B)\|} $
\end{itemize}

\subsection{AHAM heuristic}
We propose a heuristic that jointly combines the number of outliers $|k_{\text{outliers}}|$, the number of clusters $|k|$, and the similarity across the generated topic names with the $\texttt{TopicSimilarity}$ metric to the number of steps $n_{steps}$ in the domain-adaptation.  We define the \texttt{AHAM\textsubscript{objective}} as: 
$$  \texttt{AHAM\textsubscript{objective}} =  2 \cdot \frac{|k_{\text{outliers}}|}{|k|} \cdot  \sum_{\substack{i=1\\j=i+1}}^{k} \frac{\texttt{TopicSimilarity}(k_i, k_j)}{k(k - 1)} $$ We evaluate at every $10,000$ step of the GPL and select the topic modeling that yielded the lowest \texttt{AHAM\textsubscript{objective}} score within some evaluation budget of $n_{steps}$.

\section{Quantitative exploration of the AHAM objective} 
\label{sec:experimetnal}
To our knowledge, no prior research has yet investigated the impact of further domain adaptation (using methods like the aforementioned GPL) of sentence transformers within a specific domain for topic discovery in the BERTopic framework. We selected two domain datasets from the ArXiv repository \cite{muennighoff2022mteb}: \textit{arXiv}, which includes 25,000 entries from a wide range of scientific disciplines, and \textit{medarxiv}, containing 8,500 entries from the medical science field. 
The general domain dataset, \textit{arXiv}, with its larger data volume, allowed us to develop an experimental suite to assess the impact of size and domain granularity. In contrast, the smaller but domain-specific \textit{medarxiv} enabled us to examine the effects of a smaller, more specialized dataset. To address this gap, this experimental setting concentrates on four research questions:

\begin{table}[H]
    \centering
    \caption{Comparison of Topic Modeling Performance Post-Adaptation for arXiv and medarxiv Datasets Relative to Baseline, Detailing Topic Counts (\#T), Outlier Frequencies (\#O), and Outlier-to-Topic ($\frac{\#O}{\#T}$) Ratios. Bolded values represent the setting selected by the \texttt{AHAM} metric.}
   \resizebox{\textwidth}{!}{ \begin{tabular}{c| c c c c c c c |c c c c c c c}
    \toprule
    steps & domain &    \#T &  \#O & $\frac{\#O}{\#T}$ & Lev & BERT & Cos &  domain &   \#T &  \#O &  $\frac{\#O}{\#T}$ &Lev & BERT & Cos  \\ 
    \midrule
    base&      / &       15 &         43 &           2.87	&  0.32 &   0.86	 &  0.25 &      / &       15 &         43 &          2.87	&  0.32 &   0.86	 &  0.25  \\  \hline  
    
     10 &   &          20 &         32 &          1.60 &  0.32 & 0.86 & 0.31	 &  &         11 &         45 &          4.09 & 0.32 & 0.86 & 0.25 \\

    20 &   &            4 &         29 &          7.25	& 0.45 & 0.88  & 0.45  &          &    5 &         10 &          2.00 & 0.39 & 0.85  &  0.45 \\
     30 &   arXiv  &          8 &         35 &         4.38	& 0.33 & 0.87 & 0.36	 &  medarXiv  &          \textbf{11} &         \textbf{12}     &   \textbf{1.09}	& 0.32 &    0.88 & 0.41 \\
    40 & &            \textbf{19}  &          \textbf{5} &          \textbf{0.26} &  \textbf{0.31} &  \textbf{0.85}& \textbf{0.30}  &  & 4 &         23 &          5.75	& 0.29 & 0.87  & 0.42\\

    50 &    &         19 &         27 &         1.42	 & 0.35 & 0.87 &  0.38	 &   &           5 &          9 &         1.80	& 0.32 & 0.86 &0.47  \\
    \end{tabular}}
    \label{tab:adapt_anal}
\end{table}

\subsubsection{RQ1. Does the domain-specific adaptation of the sentence transformer lead to a more precise topic differentiation and a decrease in outlier topics?} 
\begin{figure}[h]
    \centering
    \includegraphics[width=\textwidth]{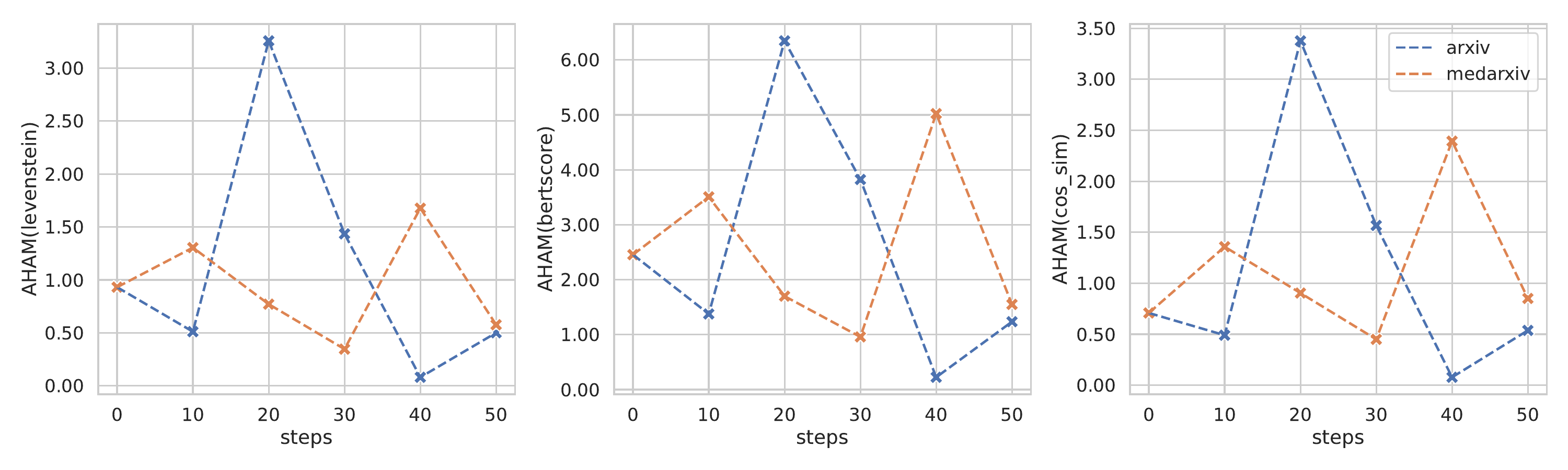}
    \caption{Assessment of the optimization objective's trajectory and similarity, conducted on intervals of 10k steps within a total budget of 50k steps.}
    \label{fig:aham-trajc}
\end{figure}

Table \ref{tab:adapt_anal}  and Figure \ref{fig:aham-trajc} provide insight into the impact of domain-specific adaptation of a sentence transformer on topic modeling within arXiv and medarxiv datasets. The results indicate a significant decrease in both the number of outliers and the number of topics at step 20k (improving from 43 outliers to 29 for the arXiv and 10 for the medarXiv), which suggests that the adaptation process likely improves the transformer's ability to discern more relevant topics and reduces the identification of outlier topics. This enhancement in precision is further evidenced by the sharp reduction in topics, pointing to a more focused topic differentiation. 

%

\subsubsection{RQ2. Which is more effective for meta-literature analysis: broad domain-specific adaptation or niche-specific knowledge?} The trajectory of the AHAM objective revealed that both lexical and semantic distance metrics peaked at step 20 for the arXiv and medarxiv datasets, indicating a two-phase process when adapting: obfuscation of knowledge and iterative refinement and improvement. We noticed that by utilizing broad domain knowledge, the adaptation process took more steps however it yielded a lower AHAM score (0.26) with 88.4\% reduction in outliers. On the other side the medarXiv adaptation, found the local optimum after 30k steps, yielding a score of 1.09 and $72\%$ reduction of outliers. The results suggest that if possible to identify what domain is best suited to transform from, then adapting such domain would enable the model to generalize better. In the case of medarXiv, we noticed that the model would specialize earlier for the topics correlated with a certain domain. 


\subsubsection{RQ3.  What is the relationship between the domain adaptation granularity and performance?} 
\begin{figure}[H]
    \centering
    \includegraphics[width=\textwidth,height=6cm]{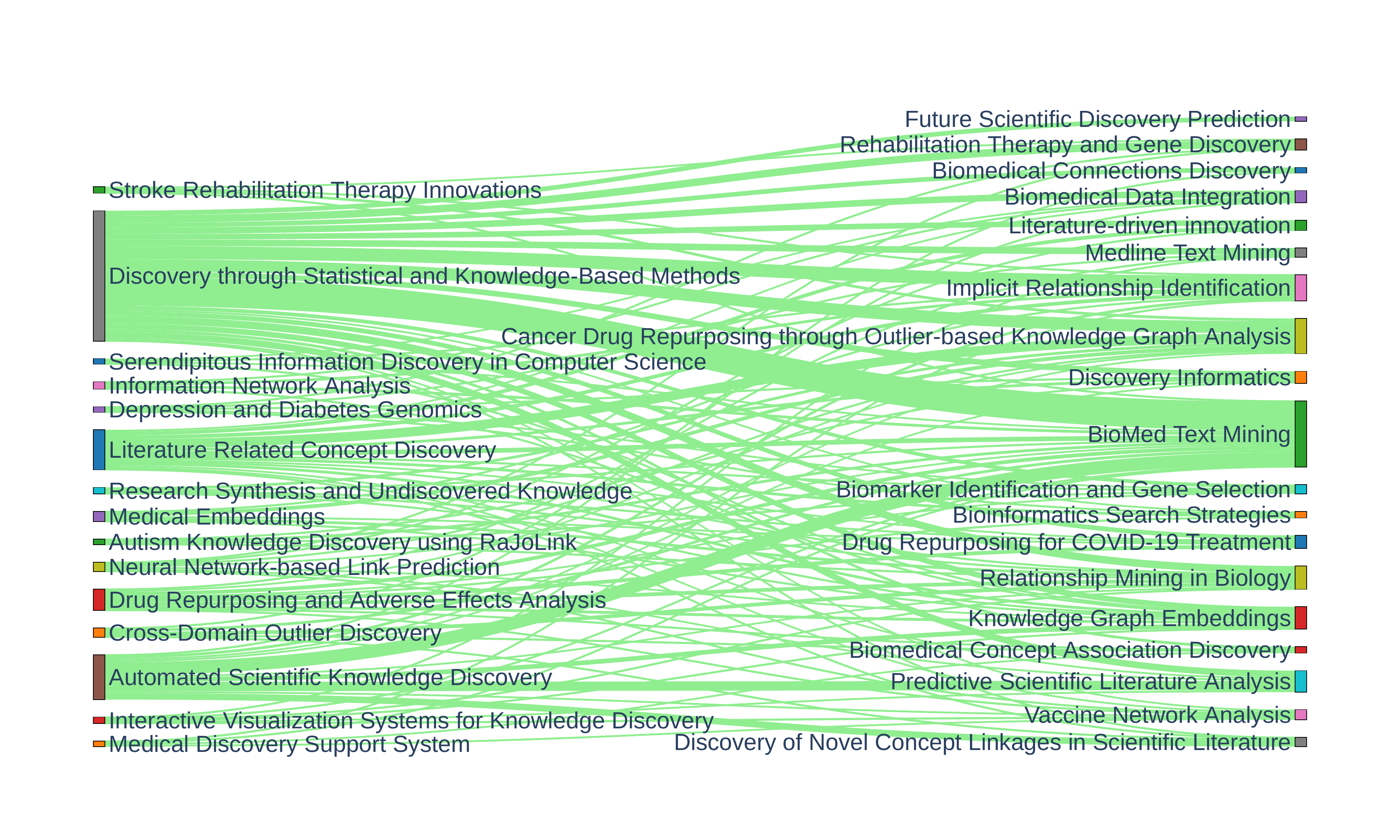}
    \caption{Evolution of topics between the first and best step of domain adaptation following AHAM objective.}
    \label{fig:enter-label2}
\end{figure}

Integrating the insights from the AHAM objective with the similarity metrics reveals that the specific stages in the modeling process significantly influence the development of topic similarities. Both datasets exhibit a tendency to diverge and then re-converge, which may reflect a common stabilization point in topic similarity as the models are refined. Evaluating the AHAM objective trajectory reveals that the function is non-convex, which makes it difficult to optimize. The results indicate a non-linear and complex progression of topic modeling characterized by both convergence and divergence, suggesting that topic evolution is influenced by the interplay of the dataset characteristics (Figure \ref{fig:enter-label2} presents the topic evolution between the initial and the topic modeling after the domain adaptation selected by AHAM). 

\subsubsection{RQ4. How do the topics evolve when the sentence-transformer language model is increasingly adapted to a specific domain?} The results of the AHAM optimization (Figure \ref{fig:aham-trajc}) suggest that extensive training with both models enhances the subsequent topic modeling's understanding of the domain it aims to represent. It is crucial to select an adaptation dataset that is closely aligned with the specific niche observed. Given the observed lexical and contextual similarities, the AHAM aspects seem to serve as reliable metrics for assessing the efficacy of the topic modeling performance of the domain-specific adaptation of the model.
\section{Conclusion and Further Work}\label{sec:fw}

In this study, we have proposed \texttt{AHAM}a metric to optimize the topic modeling via domain adaptation for meta-literature analysis and literature-based discovery by enhancing the BERTopic framework through domain adaptation. Our findings are twofold. 
\par Firstly, document partitioning has shown promising results in terms of cluster validity. Experts in the LBD manually examined three clustering solutions (i.e., baseline, ArXiv-, and medarxiv-specific adaptation models). The evaluated approaches yielded a single large general cluster, labeled ``Discovery through Statistical and Knowledge-Based Methods'' for the baseline approach, ``BioMed Text Mining'' in the ArXiv-specific model, and ``Knowledge Linkage Discovery'' for the medarxiv-specific model. This was followed by numerous smaller and highly specialized clusters, such as the ``Kostoff's cluster,'' which is heavily focused on the author's notion of literature-related discovery; the ``Biomedical Connections Discovery'' cluster, which comprises works that employ NLP-specific techniques for LBD; or the ``Knowledge Graph Embeddings'' topic, which contains recent studies utilizing link prediction on vector embeddings derived from biomedical knowledge graphs. 
\par Secondly, we have devised the AHAM heuristic based on the reduction of outliers and similarity between topics derived by LLMs by following the domain-experts prompt. This heuristic suggests a gradual, grid-like domain adaptation of the model, conditional upon a decreasing number of outliers when detecting topics. The best-defined topic modeling, yielded by lowest AHAM objective, conditioned on domain adaptation, yielded the aforementioned topics that multiple domain experts carefully inspected. The objective's optimization trajectory is complex and non-convex, suggesting that prolonged adaptation to a domain might not lead to an optimal topic modeling setting.
\par For future work, we plan to rigorously test domain adaptation in topic modeling, driven by the AHAM objective across various domains. We are also interested in exploring the effects of training with in-domain data. Additionally, an important direction is to analyze the model's capability to adapt and identify topics in different low- and less-resourced languages. An intriguing area for further investigation is evaluating the impact of the generative LLM size on topic name generation, as well as the selection of different sentence-transformers for this task. \\

\section*{Implementation} Code available at \url{https://github.com/bkolosk1/aham}.
\\
\section*{Acknowledgements}
The authors acknowledge the financial support from the Slovenian Research And Innovation Agency through research core funding (No. P2-0103) and research projects: Research collaboration prediction using literature-based discovery approach  (No. J5-2552), Embeddings-based techniques for Media Monitoring Applications (No. L2-50070) and Hate speech in contemporary conceptualizations of nationalism, racism, gender and migration (No. J5-3102).
\bibliographystyle{splncs04}
\bibliography{references}

\end{document}